\documentclass[runningheads]{llncs}

\usepackage[latin1]{inputenc}
\usepackage[english]{babel}
\usepackage{amssymb}
\usepackage{graphicx}
\usepackage{amsmath}
\usepackage{color}
\usepackage{colortbl}
\usepackage[table]{xcolor}
\usepackage{floatrow}
\usepackage{bm}
\usepackage{xspace}
\usepackage{amssymb}
\usepackage{amsmath}
\usepackage{amsfonts}
\usepackage{euscript}

\newcommand{\bby}{{\bf y}}

\newcommand{\bbw}{{\bf w}}

\newcommand{\bbX}{{\bf X}}
\newcommand{\bbZ}{{\bf Z}}

\newcommand{\bbW}{{\bf W}}

\newcommand{\mN}{{\mathcal N}}

\def\registered{{\ooalign {\hfil\raise .05ex\hbox{\scriptsize
R}\hfil\crcr\mathhexbox20D}}}

\def\REgistered{{\ooalign
{\hfil\raise.09ex\hbox{\tiny \sf R}\hfil\crcr\mathhexbox20D}}}

\makeatletter
\DeclareRobustCommand\onedot{\futurelet\@let@token\@onedot}
\def\@onedot{\ifx\@let@token.\else.\null\fi\xspace}

\definecolor{gold}{rgb}{0.85,.66,0}   

\newcommand{\revision}[1]{\textcolor{black}{#1}}

\graphicspath{{pics/}{figs/}{figs/segOverlays/}{artwork/}}

\usepackage{booktabs}
\begin{document}

\mainmatter  %

\title{Quantifying Confounding Bias in Neuroimaging Datasets with Causal Inference} %
\titlerunning{Quantifying Confounding Bias in Neuroimaging Datasets}

\author{Christian Wachinger$^{1}$, Benjamin Gutierrez Becker$^{1}$, \\Anna Rieckmann$^{2}$, Sebastian Pölsterl$^{1}$} %

\institute{$^1$Artificial Intelligence in Medical Imaging (AI-Med), KJP, LMU München \\
$^2$Department of Radiation Sciences, Ume{\aa} Univeristy }
\authorrunning{Wachinger et al.}

\maketitle

\begin{abstract}
Neuroimaging datasets keep growing in size to address increasingly complex medical questions. 
However, even the largest datasets today alone are too small for training complex
machine learning models.
A potential solution is to increase sample size by pooling scans from several datasets. 
In this work, we combine 12,207 MRI scans from 15 studies and show that
simple pooling is often ill-advised due to introducing various types of biases in
the training data.
First, we systematically \emph{define} these biases.
Second, we \emph{detect} bias by experimentally showing that scans can be
correctly assigned to their respective dataset with 73.3\% accuracy.
Finally, we propose to tell causal from confounding factors by
\emph{quantifying} the extent of confounding and causality
in a single dataset using causal inference.
We achieve this by finding the simplest graphical model in terms
of Kolmogorov complexity.
As Kolmogorov complexity is not directly computable, we employ the minimum description
length to approximate it.
We empirically show that our approach is able to estimate plausible
causal relationships from real neuroimaging data.
\end{abstract}

\section{Introduction}
As neuroimaging is joining the ranks of a ``big data'' science with more and larger datasets becoming available~\cite{smith2018statistical}, the issue of dataset bias is becoming apparent. 
Usually, neuroimaging datasets are collected with a particular research
question in mind, and inclusion criteria are tailored to answering this
particular question.
With the advancement of machine learning, in particular deep learning,
researchers require large sample sizes for training their models.
However, pooling data from studies that have been designed with
different research questions in mind, will likely lead to bias
in the learned model.
For the same reason, an estimate of a statistic from one dataset,
will likely differ from the estimate in another dataset.
In general, bias refers to an estimate of a statistic that is systematically different from its population value. 
If estimates would be truly unbiased on a population level, models would
naturally generalize to other datasets.
However, in practice, neuroimaging datasets are subject to various types of biases, including subject selection, acquisition method, and processing biases~\cite{guadalupe2017human,kruggel2010impact}.
\emph{Selection bias} stems from the fact that subjects included in the study do not represent the overall population. 
Examples are: (i) the recruitment of particular target groups, e.g., young adults; (ii) the recruitment of a particular disease group; or (iii) an over-representation of more educated participants in convenience samples. 
While the first two are potentially related to the study objective and can be controlled for, the third one is more difficult to control and also seems to appear in epidemiological studies~\cite{smith2018statistical}. 
A second bias stems from the \emph{image acquisition}, where magnetic field strength, manufacturer, gradients, pulse sequences and head positioning cause variations in the images.
While standardization efforts are undertaken for instance by the ADNI~\cite{jack2008alzheimer}, variations related to the scanner remain~\cite{kruggel2010impact}, and it is even questionable if a further standardization is in the manufacturer's interest. 
In addition, there is \emph{processing bias} in image segmentation and registration, which is in part related to acquisition bias,
because of motion artifacts, voxel sizes, and image noise that can cause
bias in segmentation results.
Moreover, different segmentation algorithms and different choices for their
hyper-parameters often lead to varying segmentation results.
Finally, an association between an image-derived measure and an outcome
is subject to \emph{confounding bias}, if both variables are affected by
a third latent random variable.

In this paper, we study bias in neuroimaging data. To this end, we combine data from 15 large-scale studies.
We propose, (i) to predict the dataset that a subject is part of as a means to detect
inter-dataset bias;
(ii) to distinguish between causal and confounded statistical relationships;
(iii) to quantify the extent of confounding and causality
in a single dataset using causal inference.
To this end, we use the algorithmic Markov condition to determine the simplest model in terms of Kolmogorov complexity to determine the true causal model.

\begin{table*}[t]
  \caption{\label{tab:dataStats}%
    Overview of neuroimaging datasets used in this study.}%
  \fontsize{8}{8}\selectfont%
  \def\arraystretch{1.3}%
  \rowcolors{2}{gray!20}{white}%
  \begin{tabular}{llrrrrrr}
  \toprule
Dataset	&	Diagnosis	&	  $N$	& \ \ \ Age (mean)  	& \ \ 	Age (SD)	& \ \ 	Males \%	 & \ \ 	Sites	& \ \	Diseased	\\
\midrule											
ABIDE I	&	Autism	&	\ \ \ \ \ 1,095	&	17.1	&	8.1	&	85.2	&	24	&	526	\\
ABIDE II	&	Autism	&	1,032	&	15.2	&	9.4	&	76.1	&	17	&	477	\\
ADHD200	&	ADHD	&	965	&	12.1	&	3.3	&	61.8	&	8	&	384	\\
ADNI	&	Alzheimer's	&	1,682	&	73.6	&	7.2	&	55.0	&	62	&	1,144	\\
AIBL	&	Alzheimer's	&	262	&	72.9	&	7.6	&	47.3	&	2	&	91	\\
COBRE	&	Schizophrenia	&	146	&	37.0	&	12.8	&	74.7	&	1	&	72	\\
CORR	&		&	1,476	&	25.9	&	15.4	&	50.0	&	32	&	0	\\
GSP	    &		&	1,563	&	21.5	&	2.8	&	42.3	&	5	&	0	\\
HBN	    &	Psychiatric	&	689	&	10.7	&	3.6	&	59.7	&	2	&	497	\\
HCP	    &		&	1,113	&	28.8	&	3.7	&	45.6	&	1	&	0	\\
IXI	    &		&	561	&	48.6	&	16.5	&	44.6	&	3	&	0	\\
MCIC	&	Schizophrenia	&	194	&	33.1	&	11.5	&	71.6	&	3	&	104	\\
NKI	    &	Psychiatric	&	624	&	38.4	&	22.5	&	39.1	&	1	&	268	\\
OASIS	&	Alzheimer's	&	415	&	52.8	&	25.1	&	38.6	&	1	&	100	\\
PPMI	&	Parkinson's	&	390	&	61.2	&	10.0	&	62.6	&	-	&	284	\\
  \bottomrule
 \end{tabular}
\end{table*}

\vspace{1ex}
\noindent
\textit{Related Work:} 
Correcting confounding factors in neuroimaging have to date been mainly studied within datasets~\cite{dukart2011age,linn2016addressing,rao2017predictive}, but not across datasets. The focus of these studies has primarily been on age and sex as confounders. 
The harmonization of cortical thickness across datasets was studied in \cite{fortin2018harmonization}. 
An instance reweighting approach for domain adaptation in Alzheimer's prediction was proposed in \cite{wachinger2016domain}. 
In contrast to prior work,
note that our proposed approach aims to detect and quantify bias
using causal inference, but not to remove bias.

\section{Data}

We work on MRI T1 brain scans from 15 large-scale public datasets: ABIDE I+II~\cite{ABIDE}, ADHD200~\cite{ADHD200}, ADNI~\cite{jack2008alzheimer}, AIBL~\cite{AIBL}, COBRE~\cite{COBRE}, CORR~\cite{zuo2014open}, GSP~\cite{GSP}, HBN~\cite{alexander2017open}, HCP~\cite{HCP}, IXI\footnote{http://brain-development.org/ixi-dataset/}, MCIC~\cite{MCIC}, NKI~\cite{nooner2012nki}, OASIS~\cite{OASIS}, and PPMI~\cite{PPMI}. 
All datasets were processed with FreeSurfer~\cite{fischl2002whole} version 5.3. 
We keep only one scan per subject from longitudinal or test-retest datasets. 
After exclusion of scans with processing errors and incomplete meta data,
scans from 12,207 subjects
(6,827 male; 8,126 controls; mean age: $33.5 \pm 23.9$) remained.
We present an overview of demographics per dataset in Table~\ref{tab:dataStats}.

\section{Name That Dataset} 

\begin{figure}[t]
\begin{center}
	\includegraphics[width=0.49\textwidth]{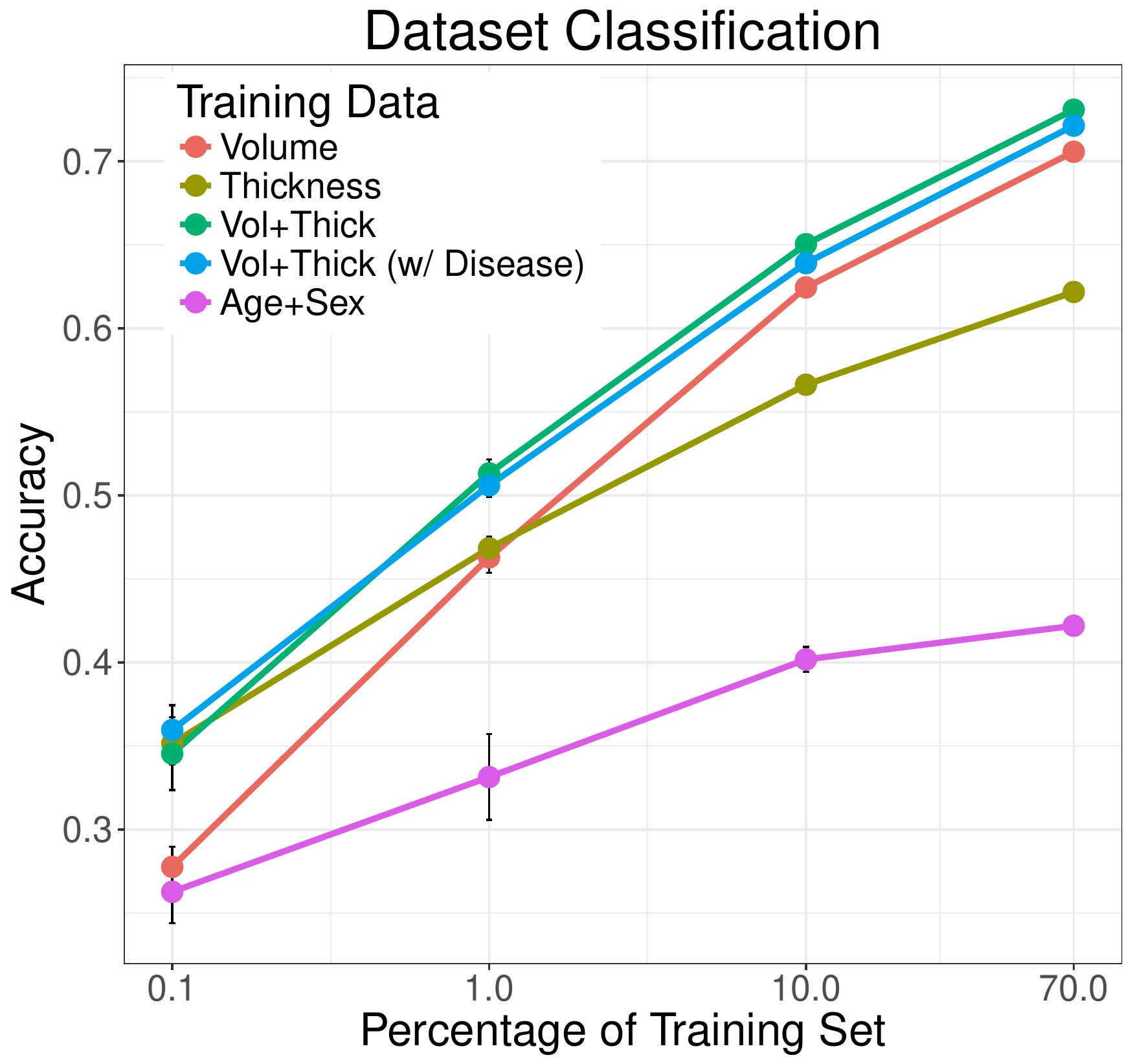} 
	\includegraphics[width=0.5\textwidth]{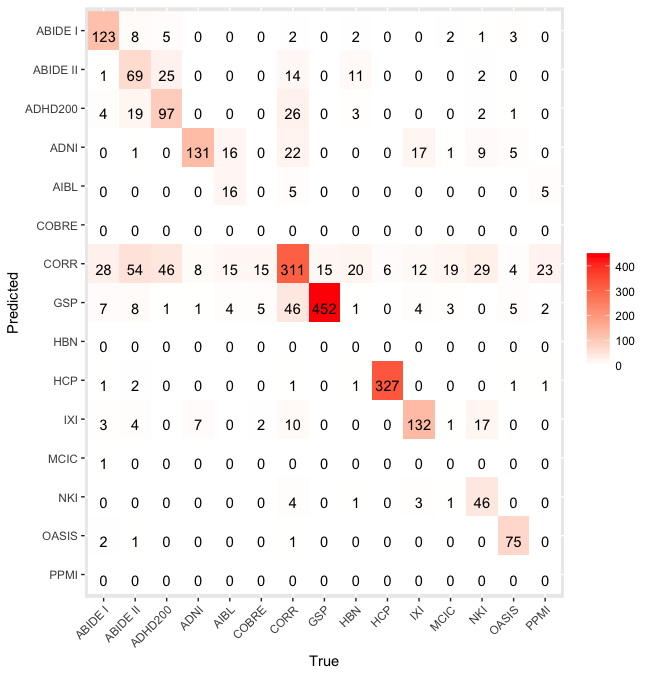} 
\caption{Left: Dataset classification accuracy for age and sex, volume, thickness, and their combination. The percentage of the data used for training is shown in log-scale. Lines show the average score over 50 repetitions, error bars show the standard deviation.
Right: Confusion matrix for volume and thickness with 70\% training data. %
\label{fig:DatasetClass}%
}
\end{center}
\end{figure}

In order to evaluate the impact of dataset bias, we  play the game \emph{Name That Dataset} on neuroimaing data that was originally proposed by Torralba and Efros~\cite{torralba2011unbiased} on natural images. 
The task is to predict the dataset an MRI scan is coming from solely based on image-derived measures, where we use volume and thickness.
We mainly focus on healthy controls to exclude disease-specific effects
that would ease classification.
Fig.~\ref{fig:DatasetClass} illustrates the performance for classifying
the 15 datasets for different combinations of image features. 
A random forest classifier with default settings was used for the prediction. %
When splitting data into training and testing sets, we take the dataset membership into account to ensure each dataset is accordingly represented.

If dataset bias would be absent, we would expect a prediction accuracy
close to random chance (6.7\% for 15 datasets).
As not all datasets have the same size and have different distributions of age and sex, we compare to results of a classifier trained on age and sex alone as baseline. 
With only 0.1\% of the data used for training, volume measures perform similar to prediction with age and sex. 
As we increase the amount of training data to 70\%, the accuracy increases to 73.3\% for the combination of volume and thickness features, which perform better than each of them alone. 
The classifier with age and sex has 42.2\% accuracy, which is well above random and therefore hints at selection bias. But compared to the 73.3\% accuracy for image features, there must be another source of bias that cannot be explained by basic demographics, such as confounding bias. 
Fig.~\ref{fig:DatasetClass} also shows similar
results after including diseased patients, denoted by `(w/ Disease)'. 

From the confusion matrix in Fig.~\ref{fig:DatasetClass}, we can see that datasets with a similar population result in higher confusion, e.g., between ABIDE I+II, and ADHD200. 
Single-site datasets, like HCP, that have strict inclusion criteria and
imaging protocols show almost no confusion with any of the other datasets.
In contrast, multi-site datasets like CORR that also cover a wide age range, show high confusion with other datasets. 
Overall, the high classification accuracy (diagonal elements) indicates that datasets possess unique, identifiable characteristics.

The lesson learned from this experiment is
that even when working with image-derived values that represent physical measures (volume, thickness), substantial bias in datasets remains, despite techniques like atlas renormalization~\cite{han2007atlas} were employed to improve consistency across scanners. 
Of course, much of the bias can be attributed to the different goals of the studies, like the inclusion of subjects from different age groups. 
However, even when focusing on datasets that cover a similar age range, we observe a high accuracy, e.g., ABIDE I and II. 
While we are not aware of previous attempts to \emph{Name That Dataset}, our results echo concerns raised in previous studies. 
In an ENIGMA study with over 15,000 subjects on brain asymmetry \cite{guadalupe2017human}, it was reported that dataset heterogeneity explained over 10\% of the total observed variance per structure. 
In ADNI, which has an optimized MPRAGE imaging protocol across all sites~\cite{jack2008alzheimer}, the intra-subject variability of compartment volumes for scans on different scanners was roughly 10 times higher than repeated scans on the same scanner~\cite{kruggel2010impact}.  
Similarly, \cite{wachinger2016domain} reported a drop in accuracy when training and testing on different datasets.

\section{Telling Causal from Confounded with Causal Inference}

\begin{figure}[t]
    \centering%
    \includegraphics[width=0.62\textwidth]{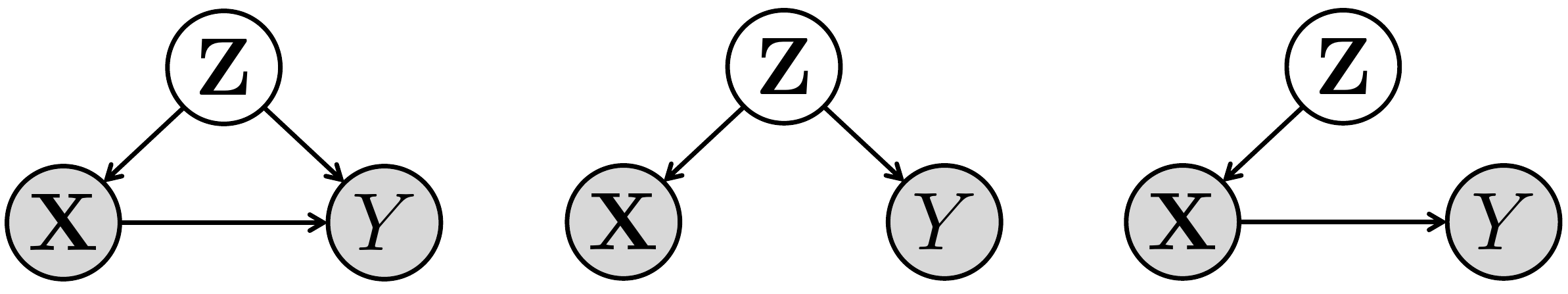}%
    \caption{Probabilistic graphical models for observed variables $\bbX$, $Y$ and unobserved confounders $\bbZ$. The statistical relationship between $\bbX$ and $Y$ is due to confounder~$\bbZ$ and due to the influence of $\bbX$ on $Y$ (left). Limiting cases are pure confounding (middle) and pure causality (right).%
    \label{fig:causalDiag}%
    }%
\end{figure}

In the previous section, we have established that there is correlation between a
feature vector $\bby$, derived from MRI scans, and the dataset $D$ the scan belongs to,
by estimating the probability $P(D=d\,|\,Y=\bby)$ via a random forest classifier.
While this has yielded useful insights, it is flawed:
(i) it cannot be used with MRI scans from a single dataset only,
and (ii) it only provides a measure of statistical dependence,
which alone is insufficient to determine causal structures of confounding bias.
Here, we want to study bias in a more principal manner by
looking at confounding bias in a causal inference framework.
Given one particular dataset, we want to quantify to what extent the correlation
between biological factors $\bbX = \{X_1,\ldots,X_m\}$ and a single image-derived measure $Y_j$
is due to $\bbX$ influencing $Y_j$, i.e., an underlying
neurobiological cause, and to what extent it is due
to other common causes, i.e., confounders $\bbZ = \{Z_1,\ldots,Z_k\}$.
In the purely causal setting, where there is no confounding, there is
a causal relationship between biological variables
$\bbX$ and image-derived measure $Y_j$,
which we denote as $\bbX \rightarrow Y_j$.
On the other end of the spectrum, the correlation between $\bbX$ and $Y_j$ is entirely
due to other measured or unmeasured causes
that have an effect on both the biological factors \emph{and} image-derived measures.
This would constitute a purely confounded relationship:
$\bbX \leftarrow \bbZ \rightarrow Y_j$.
Fig.~\ref{fig:causalDiag} illustrates the different scenarios, where
the statistical dependency between $\bbX$ and $Y_j$ is due
confounding (middle), a causal relationship (right),
or a mixture of both (left).

\subsection{Causal Inference by Minimum Description Length}

Inferring causal relations from observational data is challenging and
is only attainable under specific model assumptions.
We use an approach that incorporates the algorithmic Markov condition~\cite{kaltenpoth2019we},
which states that the simplest factorization of the joint distribution $P(X, Y)$, in terms of
Kolmogorov complexity, corresponds to the true generative process.
As Kolmogorov complexity is not directly computable, we employ the minimum description
length to approximate it.
Here, we consider two factorizations of $P(X, Y)$ as depicted in Fig. \ref{fig:causalDiag}:
(i)
the causal model $P(X, Y) = P(Y|X) P(X|Z) P(Z)$ is represented by a linear regression
model, and (ii) the confounded model $P(X, Y) = P(Y|Z) P(X|Z) P(Z)$ by probabilistic PCA. 
Thus, the complexity under the causal model can be estimated by minimum description length
$L_{\text{ca}}(\bbX, Y_j)$ via:
\begin{equation}
\begin{split}
    &L_{\text{ca}}(\bbX, Y_j) = - \log P(\bbX) \int P(Y_j | \bbX, \bbw) P(\bbw)  \mathrm{d} \bbw , \\
    &X_i \sim \mN(0, \sigma_x^2I), \qquad
    \bbw \sim \mN(0,\sigma_w^2I), \qquad
    Y_j | \bbX, \bbw \sim \mN(\bbw^\top \bbX,\sigma_y^2I) .
\end{split}
\end{equation}
The complexity of the confounded model can be estimated by
\begin{equation}
\begin{split}
    &L_{\text{co}}(\bbX,Y_j) = - \log  \int P(\bbX,Y_j | \bbZ, \bbW)  P(\bbZ) P(\bbW)  \mathrm{d} \bbW \mathrm{d} \bbZ, \\
    &Z_i \sim \mN(0, \sigma_z^2I), \qquad
    W_i \sim \mN(0,\sigma_w^2I), \qquad
    \bbX | \bbZ, \bbW \sim \mN(\bbW^\top \bbZ,\sigma_x^2I) ,
\end{split}
\end{equation}
where the confounders $\bbZ$ and the principal axes $\bbW$ are estimated using probabilistic PCA
as in \cite{kaltenpoth2019we}.
Note that we do not require that the confounders are known or measured; since
we marginalize over $\bbZ$, we only need to specify its dimensionality $k$.
To compare the causal ($\bbX \rightarrow Y_j$) and the confounded model
($\bbX \leftarrow \bbZ \rightarrow Y_j$),
we just need to compute $\Delta(\bbX, Y_j) = L_{\text{co}}(\bbX, Y_j) - L_{\text{ca}}(\bbX, Y_j)$.
If the causal model better describes the data than the confounded model,
we obtain $\Delta(\bbX, Y_j) > 0$ -- the more positive, the more confident we are.
If instead $\Delta(\bbX, Y_j) < 0$, the roles are reversed.
We estimate the complexity of both models with automatic differentiation variational inference~\cite{kucukelbir2017automatic}, efficiently implemented in PyMC3,
and use $k=1$ as the dimensionality of $\bbZ$ throughout our experiments.

\begin{figure}[t]
    \centering
    \includegraphics[scale=.323]{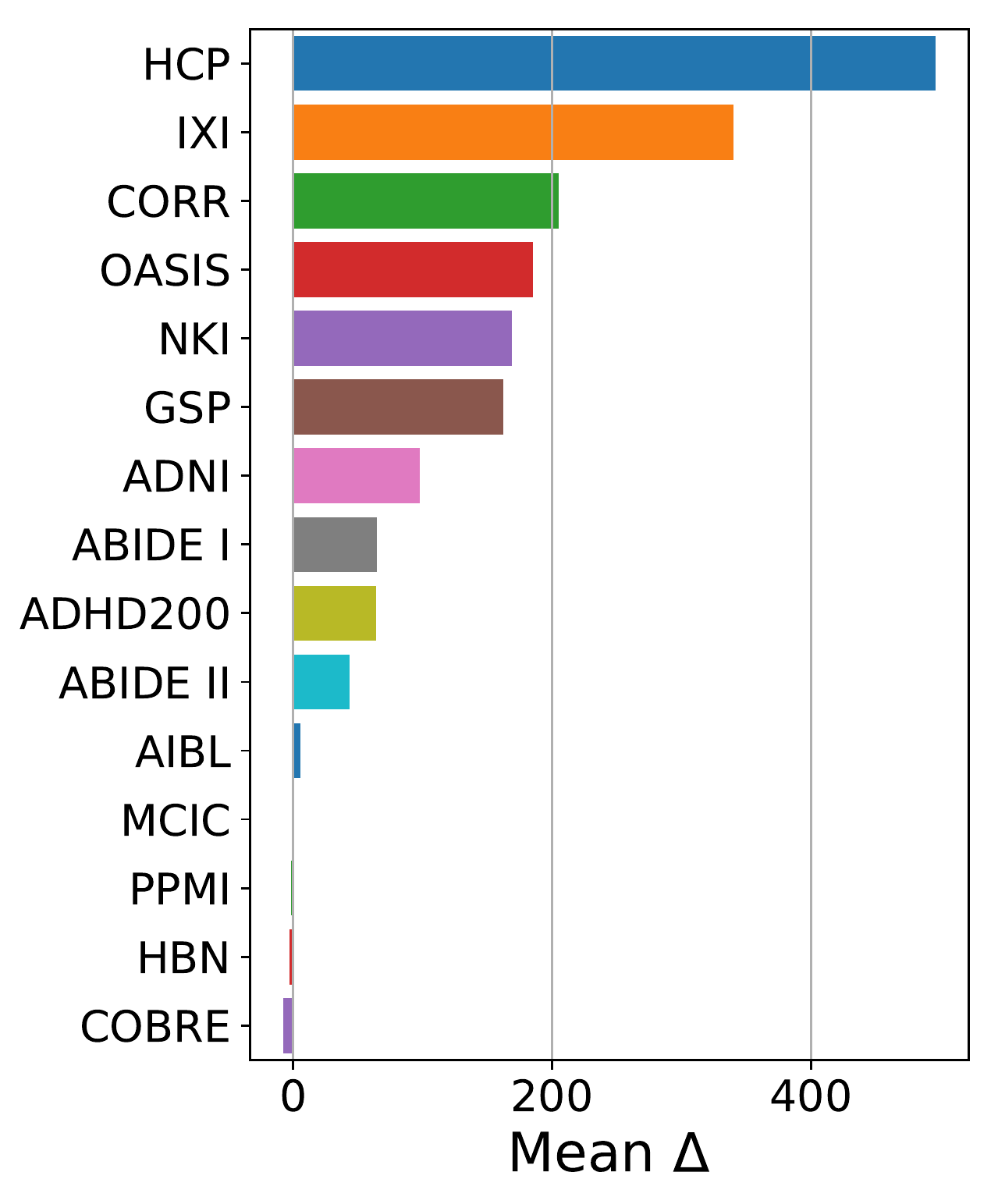}\hfill%
    \includegraphics[scale=.323]{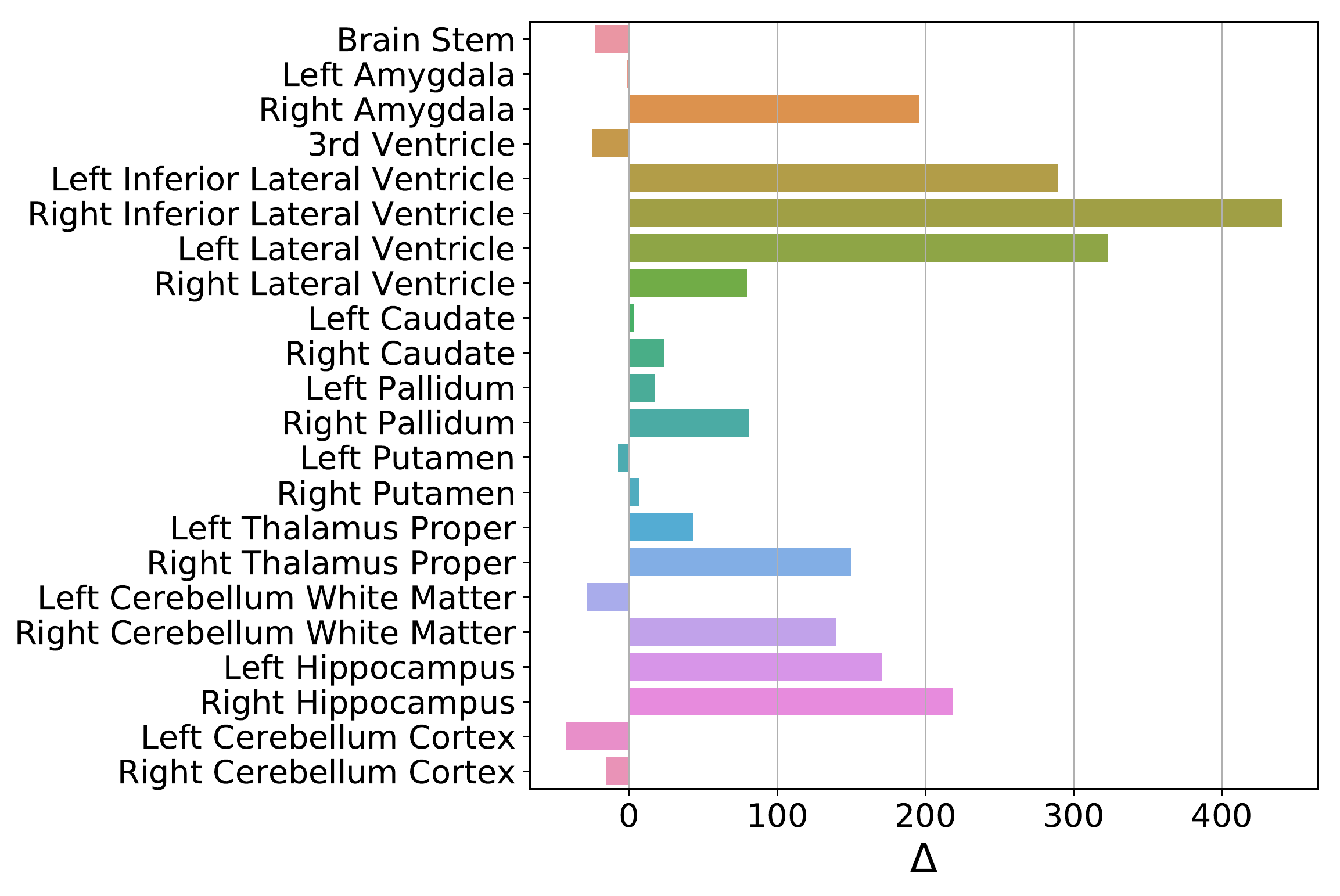}%
    \caption{Left: Mean difference $\Delta$ across brain structures for all datasets.
    Higher values indicate datasets where age and sex have a stronger causal effect on volume.
    Right: Differences $\Delta$ for all brain structures on the ADNI dataset.}%
    \label{fig:age_vs_mfs}%
\end{figure}

\subsection{Results}
Based on neurobiology, we use age, $\text{age}^2$, and sex as presumed causes ($\bbX$) of the volume of one brain structure ($Y_j$), where $j$ denotes one of 22 brain structures.
We estimate the complexity of the causal and confounded model across all brain structures and datasets, where we only select healthy subjects to facilitate comparison. 
Fig.~\ref{fig:age_vs_mfs} (left) shows the mean difference
between confounded and causal models ($\Delta$) across volumes of
all brain structures.
Positive values indicate evidence for a predominantly causal relationship
(age, sex) $\rightarrow$ volume, whereas negative values indicate a
predominantly confounded relationship.
The average relationship is most causal in HCP; or in other words,
the estimated impact of confounders is low. 
\revision{This is not surprising, considering all MRI scans in HCP have been acquired on a single, dedicated scanner that was customized for this project.}
In contrast, COBRE, HBN, PPMI, MCIC, and AIBL have zero or negative mean difference.
While we do not know the actual source of confounding, the result seems plausible,
because these datasets comprise diverse scans from various sites and scanners
(cf. table~\ref{tab:dataStats}), which likely constitute sources of
confounding. Consequently, we are unable to reliably attribute the relationship
between age, sex, and volume to either model in these datasets.
We also note that these datasets have among the lowest accuracy in the classification experiment in 
Fig.~\ref{fig:DatasetClass}.
ABIDE I+II and ADHD200 pooled previously collected data from independent sites, yielding fairly heterogeneous datasets. 
Nevertheless, their estimated causal effect is weaker compared to ADNI, GSP, and IXI, which put more effort into standardizing imaging protocols.
\revision{Harmonization across sites may strengthen the causal relationship in multi-site data.} %

In Fig.~\ref{fig:age_vs_mfs} (right), we focus on individual
differences $\Delta$ for each brain structure in the ADNI dataset,
due to its prominent role in the community.
Our estimate is most causal for lateral and inferior lateral ventricles,
which is consistent with the characteristic expansion of ventricular spaces in older ages. 
We observed the strongest confounding bias for cerebellum cortex, suggesting
the association between age, sex and cerebellum cortex is more likely
due to confounders.

\section{Conclusion}
We defined various forms of bias common to neuroimaging data.
Based on a dataset with more than 12,000 individuals,
we have demonstrated that simply pooling scans from distinct
studies can introduce substantial bias that would be passed on to
a machine learning model trained on the pooled data.
Next, we introduced a novel approach for differentiating causal from
confounded relationships based on causal inference. 
We estimated the strength of the neurobiological causal model -- age and sex influence
a brain structure's volume -- versus the confounded model -- age, sex,
and volume are influenced by latent confounders -- for 15 datasets and 22 brain structures.
Results yielded large differences in the causal strength across datasets and brain structures.
\revision{These results are specific to our assumptions for causal and confounding model, where other choices are possible and may yield different conclusions.}
Overall, we believe that the growing amount of neuroimaging data necessitates to quantify
bias in datasets and image-derived features, for which we have proposed a causal model in this work.
Finally, we note that our approach is not restricted to a
biology-derived causal model, but could also be used to estimate the
causal effect for other relationships, such as the effect of
magnetic field strength on signal-to-noise ratio.

\section*{Acknowledgements}
This research was partially supported by the Bavarian State Ministry of Science and the Arts in the framework of the Centre Digitisation.Bavaria (ZD.B).

\bibliographystyle{splncs03}
\bibliography{jab_bib.bib}

\end{document}